\documentclass{article}

\usepackage[final]{include/drlw_neurips_2019}

\usepackage[utf8]{inputenc} %
\usepackage[T1]{fontenc}    %
\usepackage{hyperref}       %
\usepackage{url}            %
\usepackage{booktabs}       %
\usepackage{amsfonts}       %
\usepackage{nicefrac}       %
\usepackage{microtype}      %
\usepackage{array}

\usepackage{microtype}
\usepackage{graphicx}
\usepackage{subfigure}
\usepackage{caption}
\usepackage{amsmath}
\usepackage{wrapfig}

\renewcommand{\cite}{\citep}

\title{The StarCraft Multi-Agent Challenge}

\usepackage[]{authblk}

\makeatletter
\renewcommand\AB@affilsepx{ ~ \protect\Affilfont}
\makeatother

\renewcommand*{\Affilfont}{\normalsize\normalfont}

\author[1]{Mikayel~Samvelyan$^*$}
\author[2]{Tabish~Rashid$^*$}
\author[2]{Christian~Schroeder~de~Witt}
\author[2]{Gregory~Farquhar}
\author[2]{Nantas~Nardelli}
\author[2]{Tim~G.~J.~Rudner}
\author[2]{Chia-Man~Hung}
\author[2]{Philip~H.~S.~Torr}
\author[3]{Jakob~Foerster}
\author[2]{Shimon~Whiteson}
\affil[1]{Russian-Armenian University}
\affil[2]{University of Oxford}
\affil[3]{Facebook AI Research}
\newcommand{\customfootnotetext}[2]{{%
		\renewcommand{\thefootnote}{#1}%
		\footnotetext[0]{#2}}}%

\begin{document}

\maketitle
\vspace*{-0.5cm}
\customfootnotetext{$^*$}{Equal contribution. Correspondence to \href{mailto:mikayel@samvelyan.com}{<mikayel@samvelyan.com>} or \href{mailto:tabish.rashid@cs.ox.ac.uk}{<tabish.rashid@cs.ox.ac.uk>}.}

\begin{abstract}
In the last few years, deep multi-agent reinforcement learning (RL) has become a highly active area of research. A particularly challenging class of problems in this area is partially observable, cooperative, multi-agent learning, in which teams of agents must learn to coordinate their behaviour while conditioning only on their private observations. This is an attractive research area since such problems are relevant to a large number of real-world systems and are also more amenable to evaluation than general-sum problems.
	
Standardised environments such as the ALE and MuJoCo have allowed single-agent RL to move beyond toy domains, such as grid worlds. However, there is no comparable benchmark for cooperative multi-agent RL. As a result, most papers in this field use one-off toy problems, making it difficult to measure real progress. In this paper, we propose the StarCraft Multi-Agent Challenge (SMAC) as a benchmark problem to fill this gap.\footnote{Code is available at \url{https://github.com/oxwhirl/smac}} SMAC is based on the popular real-time strategy game StarCraft II and focuses on micromanagement challenges where each unit is controlled by an independent agent that must act based on local observations. We offer a diverse set of challenge scenarios and recommendations for best practices in benchmarking and evaluations. We also open-source a deep multi-agent RL learning framework including state-of-the-art algorithms.\footnote{Code is available at \url{https://github.com/oxwhirl/pymarl}} We believe that SMAC can provide a standard benchmark environment for years to come. Videos of our best agents for several SMAC scenarios are available at: \url{https://youtu.be/VZ7zmQ_obZ0}.
\end{abstract}
\section{Introduction}

Deep reinforcement learning (RL) promises a scalable approach to solving arbitrary sequential decision-making problems, demanding only that a user must specify a reward function that expresses the desired behaviour.
However, many real-world problems that might be tackled by RL are inherently multi-agent in nature.
For example, the coordination of self-driving cars, autonomous drones, and other multi-robot systems are becoming increasingly critical. Network traffic routing, distributed sensing, energy distribution, and other logistical problems are also inherently multi-agent. 
As such, it is essential to develop multi-agent RL (MARL) solutions that can handle decentralisation constraints and deal with the exponentially growing joint action space of many agents.

Partially observable, cooperative, multi-agent learning problems are of particular interest.
Cooperative problems avoid difficulties in evaluation inherent with general-sum games (e.g., which opponents are evaluated against). Cooperative problems also map well to a large class of critical problems where a single user that manages a distributed system can specify the overall goal, e.g., minimising traffic or other inefficiencies.
Most real-world problems depend on inputs from noisy or limited sensors, so partial observability must also be dealt with effectively. This often includes limitations on communication that result in a need for decentralised execution of learned policies.
However, there commonly is access to additional information during training, which may be carried out in controlled conditions or simulation.

A growing number of recent works \cite{foerster_counterfactual_2017, rashid_qmix:_2018, sunehag_value-decomposition_2017, lowe_multi-agent_2017} have begun to address the problems in this space.
However, there is a clear lack of standardised benchmarks for research and evaluation.
Instead, researchers often propose one-off environments which can be overly simple or tuned to the proposed algorithms.
In single-agent RL, standard environments such as the Arcade Learning Environment \cite{bellemare13arcade}, or MuJoCo for continuous control \cite{Plappert2019multigoal}, have enabled great progress.
In this paper, we aim to follow this successful model by offering challenging standard benchmarks for deep MARL and to facilitate more rigorous experimental methodology across the field.

Some testbeds have emerged for other multi-agent regimes, such as Poker \cite{HeinrichS16}, Pong \cite{tampuu_multiagent_2015}, Keepaway Soccer \cite{stone2005keepaway}, or simple gridworld-like environments \cite{lowe_multi-agent_2017, leibo_multi-agent_2017, yang2018mean, zheng2017magent}.
Nonetheless, we identify a clear gap in challenging and standardised testbeds for the important set of domains described above.

To fill this gap, we introduce the StarCraft Multi-Agent Challenge (SMAC).
SMAC is built on the popular real-time strategy game StarCraft II\footnote{StarCraft II is the sequel to the game StarCraft and its expansion set Brood War. StarCraft and StarCraft II are trademarks of Blizzard Entertainment\textsuperscript{TM}.} and makes use of the SC2LE environment \cite{vinyals_starcraft_2017}.
Instead of tackling the full game of StarCraft with centralised control, we focus on decentralised micromanagement challenges (Figure~\ref{fig:SC2maps_2}).
In these challenges, each of our units is controlled by an independent, learning agent that has to act based only on local observations, while the opponent's units are controlled by the hand-coded built-in StarCraft II AI.
We offer a diverse set of scenarios that challenge algorithms to handle high-dimensional 
inputs and partial observability, and to learn coordinated 
behaviour even when restricted to fully decentralised execution.

The full games of StarCraft: BroodWar and StarCraft II have already been used as RL environments, due to the many interesting challenges inherent to the games \cite{synnaeve_torchcraft:_2016, vinyals_starcraft_2017}. DeepMind's AlphaStar \cite{alphastar} has recently shown an impressive level of play on a StarCraft II matchup using a centralised controller. In contrast, SMAC is not intended as an environment to train agents for use in full StarCraft II gameplay. Instead, by introducing strict decentralisation and local partial observability, we use the StarCraft II game engine to build a new set of rich cooperative multi-agent problems that bring unique challenges, such as the nonstationarity of learning \cite{foerster_stabilising_2017}, multi-agent credit assignment \cite{foerster_counterfactual_2017}, and the difficulty of representing the value of joint actions \cite{rashid_qmix:_2018}.

To further facilitate research in this field, we also open-source PyMARL, a learning framework that can serve as a starting point for other researchers and includes implementations of several key MARL algorithms. PyMARL is modular, extensible, built on PyTorch, and serves as a template for dealing with some of the unique challenges of deep MARL in practice. We include results on our full set of SMAC environments using QMIX \cite{rashid_qmix:_2018} and several baseline algorithms, and challenge the community to make progress on difficult environments in which good performance has remained out of reach so far. 
We also offer a set of guidelines for best practices in evaluations using our benchmark, including the reporting of standardised performance metrics, sample efficiency, and computational requirements (see Appendix~\ref{appendix:methodology}).

We hope SMAC will serve as a valuable standard benchmark, enabling systematic and robust progress in deep MARL for years to come.

\section{Related Work}

Much work has gone into designing environments to test and develop MARL agents. However, not many of these focused on providing a qualitatively
challenging environment that would provide together elements of partial
observability, challenging dynamics, and
high-dimensional observation spaces.

\citet{stone2005keepaway} presented Keepaway soccer, a domain built on the RoboCup soccer
simulator \citep{kitano1997robocup}, a 2D simulation of a football environment
with simplified physics, where the main task consists of keeping a ball within a
pre-defined area where agents in teams can reach, steal, and pass the ball,
providing a simplified setup for studying cooperative MARL.
This domain was later extended to the Half Field Offense task
\citep{kalyanakrishnan2006half, hausknecht_half_2016}, which increases the
difficulty of the problem by requiring the agents to not only keep the ball
within bounds but also to score a goal.
Neither task scales well in difficulty with the number of agents, as most
agents need to do little coordination. There is also a lack of interesting environment
dynamics beyond the simple 2D physics nor good reward signals, thus reducing the
impact of the environment as a testbed.

Multiple gridworld-like environments have also been explored.
\citet{lowe_multi-agent_2017} released a set of simple grid-world like
environments for multi-agent RL alongside an implementation of MADDPG, featuring
a mix of competitive and cooperative tasks focused on shared communication and
low level continuous control.
\citet{leibo_multi-agent_2017} show several mixed-cooperative Markov
environment focused on testing social dilemmas, however, they did not release an
implementation to further explore the tasks.
\citet{yang2018mean, zheng2017magent} present a framework for creating
gridworlds focuses on many-agents tasks, where the number of agents ranges from
the hundreds to the millions. This work, however, focuses on testing for emergent
behaviour, since environment dynamics and control space need to remain
relatively simple for the tasks to be tractable.
\citet{resnick2018pommerman} propose a multi-agent environment based on the game
\emph{Bomberman}, encompassing a series of cooperative and adversarial tasks
meant to provide a more challenging set of tasks with a relatively
straightforward 2D state observation and simple grid-world-like action spaces.

Learning to play StarCraft games also has been investigated in several
communities: work ranging from evolutionary algorithms to tabular
RL applied has shown that the game is an excellent testbed
for both modelling and planning \citep{ontanon2013survey}, however, most have
focused on single-agent settings with multiple controllers and classical
algorithms.
More recently, progress has been made on developing frameworks that enable
researchers working with deep neural networks to test recent algorithms on these
games; work on applying deep RL algorithms to single-agent and multi-agent
versions of the micromanagement tasks has thus been steadily appearing
\citep{usunier2016episodic, foerster_stabilising_2017,
foerster_counterfactual_2017, rashid_qmix:_2018, nardelli2018value,
hu2018knowledge, shao2018starcraft, foerster2018multi} with the release of
TorchCraft \citep{synnaeve_torchcraft:_2016} and SC2LE
\citep{vinyals_starcraft_2017}, interfaces to respectively \emph{StarCraft:
BroodWar} and \emph{StarCraft II}. Our work presents the first standardised
testbed for decentralised control in this space.

Other work focuses on playing the full game of StarCraft, including 
macromanagement and tactics 
\citep{pang2018reinforcement,sun2018tstarbots,alphastar}. 
By introducing decentralisation and local observability, our agents 
are excessively restricted compared to normal full gameplay. SMAC is therefore 
not intended as an environment to train agents for use in full gameplay. 
Instead, we use the StarCraft II game engine to build rich and interesting 
multi-agent problems.
\section{Multi-Agent Reinforcement Learning}

In SMAC, we focus on tasks where a team of agents needs to work together to achieve a common goal. We briefly review the formalism of such \textit{fully cooperative multi-agent tasks} as \textit{Dec-POMDPs} but refer readers to \citet{oliehoek_concise_2016} for a more complete picture.
\paragraph*{Dec-POMDPs} Formally, a Dec-POMDP $G$ is given by a tuple 
$G=\left\langle S,U,P,r,Z,O,n,\gamma\right\rangle$, where  $s \in S$ is the true state of the environment. At each time step, each agent $a \in A \equiv \{1,...,n\}$ chooses an action $u^a\in U$, forming a joint action $\mathbf{u}\in\mathbf{U}\equiv U^n$. 
This causes a transition of the environment according to the state transition function $P(s'|s,\mathbf{u}):S\times\mathbf{U}\times S\rightarrow [0,1]$. 

In contrast to partially-observable stochastic games, all agents in a Dec-POMDP share the same team reward function $r(s,\mathbf{u}):S\times\mathbf{U}\rightarrow\mathbb{R}$, where $\gamma\in[0,1)$ is the discount factor. 
Dec-POMDPs consider \textit{partially observable} scenarios in which an observation function $O(s,a):S\times A\rightarrow Z$ determines the observations $z^a\in Z$ that each agent draws individually at each time step. 
Each agent has an action-observation history $\tau^a\in T\equiv(Z\times U)^*$, on which it conditions a stochastic policy $\pi^a(u^a|\tau^a):T\times U\rightarrow [0,1]$. The joint policy $\pi$ admits a joint \textit{action-value function}: $Q^\pi(s_t, \mathbf{u}_t)=\mathbb{E}_{s_{t+1:\infty},\mathbf{u}_{t+1:\infty}} \left[R_t|s_t,\mathbf{u}_t\right],$ where $R_t=\sum^{\infty}_{i=0}\gamma^ir_{t+i}$ is the \textit{discounted return}.

\paragraph*{Centralised training with decentralised execution}
If learning proceeds in simulation, as is done in StarCraft II, or in a laboratory, then it can usually be conducted in a centralised fashion. This gives rise to the paradigm of \textit{centralised training with decentralised execution}, which has been well-studied in the planning community \cite{oliehoek_optimal_2008,kraemer_multi-agent_2016}. Although training is centralised, execution is decentralised, i.e., the 
learning algorithm has access to all local action-observation histories 
$\boldsymbol{\tau}$ and global state $s$, but each agent's learnt policy can 
condition only on its own action-observation history $\tau^a$. 

A number of recent state-of-the-art algorithms use extra state information available in centralised training to speed up the learning of decentralised policies. Among these, COMA \cite{foerster_counterfactual_2017} is an actor-critic algorithm with a special multi-agent critic baseline, and QMIX \cite{rashid_qmix:_2018} belongs to the $Q$-learning family.

\section{SMAC}

SMAC is based on the popular real-time strategy (RTS) game StarCraft II.
In a regular full game of StarCraft II, one or more humans compete against each other or against a built-in game AI to gather resources, construct buildings, and build armies of units to defeat their opponents.

Akin to most RTSs, StarCraft has two main gameplay components: macromanagement and micromanagement. \emph{Macromanagement} refers to high-level strategic considerations, such as economy and resource management.
\emph{Micromanagement} (micro), on the other hand, refers to fine-grained control of individual units.

StarCraft has been used as a research platform for AI, and more recently, RL. Typically, the game is framed as a competitive problem: an agent takes the role of a human player, making macromanagement decisions and performing micromanagement as a puppeteer that issues orders to individual units from a centralised controller.

In order to build a rich multi-agent testbed, we instead focus solely on micromanagement.
Micro is a vital aspect of StarCraft gameplay with a high skill ceiling, and is practiced in isolation by amateur and professional players.
For SMAC, we leverage the natural multi-agent structure of micromanagement by proposing a modified version of the problem designed specifically for decentralised control.
In particular, we require that each unit be controlled by an independent agent that conditions only on local observations restricted to a limited field of view centred on that unit (see Figure \ref{fig:obs}). 
Groups of these agents must be trained to solve challenging combat scenarios, battling an opposing army under the centralised control of the game's built-in scripted AI.

Proper micro of units during battles maximises the damage dealt to enemy units while minimising damage received, and requires a range of skills.
For example, one important technique is \textit{focus fire}, i.e., ordering units to jointly attack and kill enemy units one after another. When focusing fire, it is important to avoid \textit{overkill}: inflicting more damage to units than is necessary to kill them.

\begin{figure}[t!]
	\centering
	\subfigure[\texttt{2c\_vs\_64zg}]{
		\includegraphics[width=0.32\columnwidth]{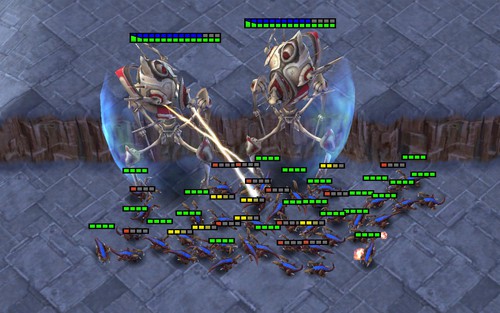}}
	\subfigure[\texttt{3s\_vs\_5z}]{
		\includegraphics[width=0.32\columnwidth]{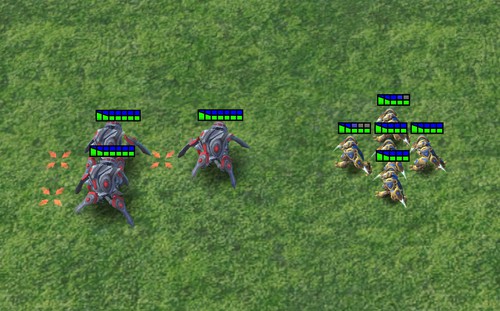}}
	\subfigure[\texttt{corridor}]{
		\includegraphics[width=0.32\columnwidth]{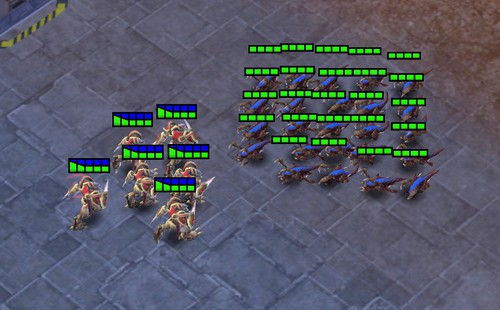}}
	\caption{\label{fig:SC2maps_2}Screenshots of three SMAC scenarios.}
		\vspace{-.5cm}
\end{figure}

Other common micro techniques include: assembling units into formations based on their armour types, making enemy units give chase while maintaining enough distance so that little or no damage is incurred (\textit{kiting}), coordinating the positioning of units to attack from different directions or taking advantage of the terrain to defeat the enemy. Figure~\ref{fig:SC2maps_2}b illustrates how three Stalker units kite five enemy Zealots units.

SMAC thus provides a convenient environment for evaluating the effectiveness of MARL algorithms. The simulated StarCraft II environment and carefully designed scenarios require learning rich cooperative behaviours under partial observability, which is a challenging task. The simulated environment also provides an additional state information during training, such as information on all the units on the entire map. This is crucial for facilitating algorithms to take full advantage of the centralised training regime and assessing all aspects of MARL methods.
SMAC is a qualitatively challenging environment that provides together elements of partial observability, challenging dynamics, and high-dimensional observation spaces.

\subsection*{Scenarios}

SMAC consists of a set of StarCraft II micro scenarios which aim to evaluate how well independent agents are able to learn coordination to solve complex tasks. 
These scenarios are carefully designed to necessitate the learning of one or more micromanagement techniques to defeat the enemy.
Each scenario is a confrontation between two armies of units.
The initial position, number, and type of units in each army varies from scenario to scenario, as does the presence or absence of elevated or impassable terrain. Figures \ref{fig:SC2maps_2} include screenshots of several SMAC micro scenarios.

The first army is controlled by the learned allied agents.
The second army consists of enemy units controlled by the built-in game AI, which uses carefully handcrafted non-learned heuristics.
At the beginning of each episode, the game AI instructs its units to attack the allied agents using its scripted strategies.
An episode ends when all units of either army have died or when a pre-specified time limit is reached (in which case the game is counted as a defeat for the allied agents).
The goal is to maximise the win rate, i.e., the ratio of games won to games played.

The complete list of challenges is presented in Table~\ref{tab:scenario}\footnote{The list of SMAC scenarios has been updated from the earlier version. All scenarios, however, are still available in the repository.}. More specifics on the SMAC scenarios and environment settings can be found in Appendices \ref{appendix:SMAC_scenarios} and \ref{appendix:SMAC_evn_setting} respectively.

\begin{table*}
	\caption{SMAC challenges.}
	\scalebox{.9}{
		\begin{tabular}{ccc}
			\toprule
			Name & Ally Units & Enemy Units \\
			\midrule
			\texttt{2s3z} &  2 Stalkers \& 3 Zealots &  2 Stalkers \& 3 Zealots \\
			\texttt{3s5z} &  3 Stalkers \&  5 Zealots &  3 Stalkers \&  5 Zealots \\
			\texttt{1c3s5z} &  1 Colossus, 3 Stalkers \&  5 Zealots &  1 Colossus, 3 Stalkers \&  5 Zealots \\
			\hline
			\texttt{5m\_vs\_6m} & 5 Marines & 6 Marines \\
			\texttt{10m\_vs\_11m} & 10 Marines & 11 Marines \\
			\texttt{27m\_vs\_30m} & 27 Marines & 30 Marines \\
			\texttt{3s5z\_vs\_3s6z} & 3 Stalkers \& 5 Zealots & 3 Stalkers \& 6 Zealots \\
			\texttt{MMM2} &  1 Medivac, 2 Marauders \& 7 Marines
			&  1 Medivac, 3 Marauders \& 8 Marines  \\
			\hline
			\texttt{2s\_vs\_1sc}& 2 Stalkers  & 1 Spine Crawler \\
			\texttt{3s\_vs\_5z} & 3 Stalkers & 5 Zealots \\
			\texttt{6h\_vs\_8z} & 6 Hydralisks  & 8 Zealots \\
			\texttt{bane\_vs\_bane} & 20 Zerglings \& 4 Banelings  & 20 Zerglings \& 4 Banelings \\		
			\texttt{2c\_vs\_64zg}& 2 Colossi  & 64 Zerglings \\		
			\texttt{corridor} & 6 Zealots  & 24 Zerglings \\
			\bottomrule
	\end{tabular}}
	\label{tab:scenario}
	\vspace{-.3cm}
\end{table*}

\subsection*{State and Observations}\label{section:state_and_obs}
\begin{wrapfigure}{l}{.45\textwidth}
	\vspace{-.5cm}
	\centering
	\includegraphics[width=\linewidth]{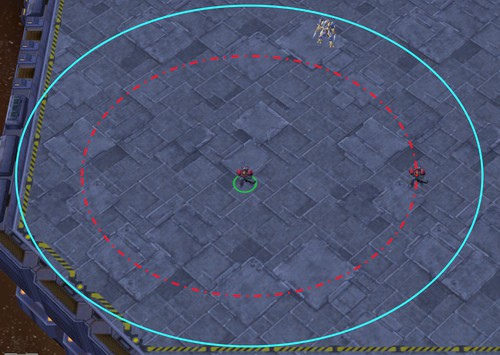}
	\captionof{figure}{The cyan and red circles respectively border the sight and shooting range of the agent.}
	\label{fig:obs}
	\vspace{-1cm}
\end{wrapfigure}

At each timestep, agents receive local observations drawn within their field of view. This encompasses information about the map within a circular area around each unit and with a radius equal to the \textit{sight range} (Figure \ref{fig:obs}). The sight range makes the environment partially observable from the standpoint of each agent. Agents can only observe other agents if they are both alive and located within the sight range. Hence, there is no way for agents to distinguish between teammates that are far away from those that are dead.

The feature vector observed by each agent contains the following attributes for both allied and enemy units within the sight range: \texttt{distance}, \texttt{relative x}, \texttt{relative y}, \texttt{health}, \texttt{shield}, and \texttt{unit\_type}. Shields serve as an additional source of protection that needs to be removed before any damage can be done to the health of units.
All Protos units have shields, which can regenerate if no new damage is dealt.
In addition, agents have access to the last actions of allied units that are in the field of view. Lastly, agents can observe the terrain features surrounding them, in particular, the values of eight points at a fixed radius indicating height and walkability.

The global state, which is only available to agents during centralised training, contains information about all units on the map. Specifically, the state vector includes the coordinates of all agents relative to the centre of the map, together with unit features present in the observations. Additionally, the state stores the \texttt{energy} of Medivacs and \texttt{cooldown} of the rest of the allied units, which represents the minimum delay between attacks. Finally, the last actions of all agents are attached to the central state.

All features, both in the state as well as in the observations of individual agents, are normalised by their maximum values. The sight range is set to nine for all agents.

\subsection*{Action Space}

The discrete set of actions that agents are allowed to take consists of \texttt{move[direction]}\footnote{Four directions: north, south, east, or west.}, \texttt{attack[enemy\_id]}, \texttt{stop} and \texttt{no-op}.\footnote{Dead agents can only take \texttt{no-op} action while live agents cannot.}
As healer units, Medivacs use \texttt{heal[agent\_id]} actions instead of \texttt{attack[enemy\_id]}. The maximum number of actions an agent can take ranges between 7 and 70, depending on the scenario.

To ensure decentralisation of the task, agents can use the \texttt{attack[enemy\_id]} action only on enemies in their \textit{shooting range} (Figure \ref{fig:obs}).
This additionally constrains the ability of the units to use the built-in \emph{attack-move} macro-actions on the enemies that are far away. We set the shooting range equal to $6$ for all agents. Having a larger sight range than a shooting range forces agents to make use of the move commands before starting to fire.

\subsection*{Rewards}

The overall goal is to maximise the win rate for each battle scenario.
The default setting is to use the \textit{shaped reward}, which produces a reward based on the hit-point damage dealt and enemy units killed, together with a special bonus for winning the battle.
The exact values and scales for each of these events can be configured using a range of flags. 
To produce fair comparisions we encourage using this default reward function for all scenarios.
We also provide another \textit{sparse reward} option, in which the reward is +1 for winning and -1 for losing an episode.

\section{PyMARL}
\label{sec:pymarl}
To make it easier to develop algorithms for SMAC, we have also open-sourced our software engineering framework PyMARL. PyMARL has been designed explicitly with deep MARL in mind and allows for out-of-the-box experimentation and development.

PyMARL's codebase is organized in a modular fashion in order to enable the rapid development of new algorithms, as well as providing implementations of current deep MARL algorithms to benchmark against. It is built on top of PyTorch to facilitate the fast execution and training of deep neural networks, and take advantage of the rich ecosystem built around it. PyMARL's modularity makes it easy to extend, and components can be readily isolated for testing purposes.

Since the implementation and development of deep MARL algorithms come with a number of additional challenges beyond those posed by single-agent deep RL, it is crucial to have simple and understandable code. In order to improve the readability of code and simplify the handling of data between components, PyMARL encapsulates all data stored in the buffer within an easy to use data structure. This encapsulation provides a cleaner interface for the necessary handling of data in deep MARL algorithms, whilst not obstructing the manipulation of the underlying PyTorch Tensors. In addition, PyMARL aims to maximise the batching of data when performing inference or learning so as to provide significant speed-ups over more naive implementations. 

PyMARL features implementations of the following algorithms: QMIX \cite{rashid_qmix:_2018}, QTRAN \cite{son2019qtran} and COMA \cite{foerster_counterfactual_2017} as state-of-the-art methods, and VDN \cite{sunehag_value-decomposition_2017} and IQL \cite{tan_multi-agent_1993} as baselines.

\section{Results}
In this section, we present results for scenarios included as part of SMAC. The purpose of these results is to demonstrate the performance of the current state-of-the-art methods in our chosen MARL paradigm.

The evaluation procedure is similar to the one in \cite{rashid_qmix:_2018}. The training is paused after every $10000$ timesteps during which $32$ test episodes are run with agents performing action selection greedily in a decentralised fashion. The percentage of episodes where the agents defeat all enemy units within the permitted time limit is referred to as the \textit{test win rate}.

The architectures and training details are presented in Appendix~\ref{appendix:exp_setup}.
A table of the results is included in Appendix~\ref{sec:table_results}.
Data from the individual runs are available at the SMAC repo (\url{https://github.com/oxwhirl/smac}).

\begin{figure*}[h!]
	\centering
	\includegraphics[width=0.45\textwidth]{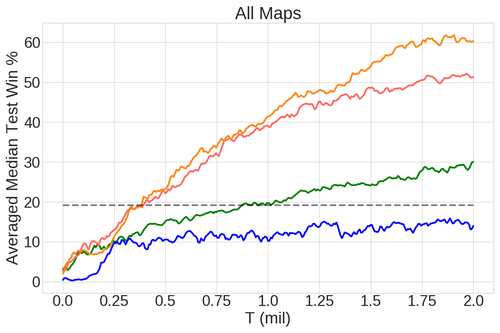}
	\includegraphics[width=0.45\textwidth]{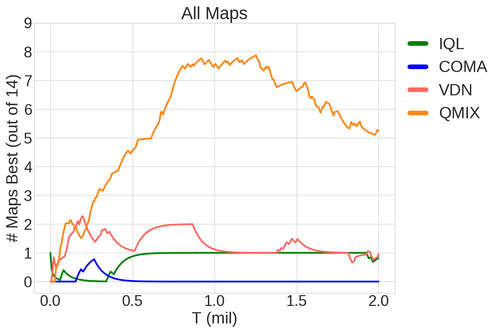}
	\caption{Left: The median test win \%, averaged across all 14 scenarios. Heuristic's performance is shown as a dotted line. Right: The number of scenarios in which the algorithm's median test win \% is the highest by at least $1/32$ (smoothed).}
	\label{fig:median_test_win}
\end{figure*}

Figure \ref{fig:median_test_win} plots the median test win percentage averaged across all scenarios to compare the algorithms across the entire SMAC suite. 
We also plot the performance of a simple heuristic AI that selects the closest enemy unit (ignoring partial observability) and attacks it with the entire team until it is dead, upon which the next closest enemy unit is selected. This is a basic form of \textit{focus-firing}, which is a crucial tactic for achieving good performance in micromanagement scenarios.
The relatively poor performance of the heuristic AI shows that the suite of SMAC scenarios requires more complex behaviour than naively focus-firing the closest enemy, making it an interesting and challenging benchmark.

Overall QMIX achieves the highest test win percentage and is the best performer on up to eight scenarios during training. 
Additionally, IQL, VDN, and QMIX all significantly outperform COMA, demonstrating the sample efficiency of off-policy value-based methods over on-policy policy gradient methods. 

Based on the results, we broadly group the scenarios into 3 categories: 
\textit{Easy},
\textit{Hard}, and
\textit{Super-Hard} based on the performance of the algorithms.

\begin{figure*}[h!]
	\centering
	\includegraphics[width=0.32\textwidth]{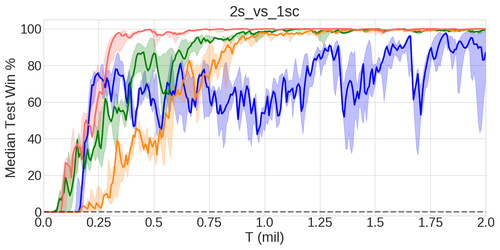}
	\includegraphics[width=0.32\textwidth]{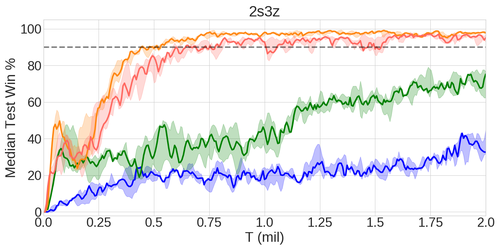}
	\includegraphics[width=0.32\textwidth]{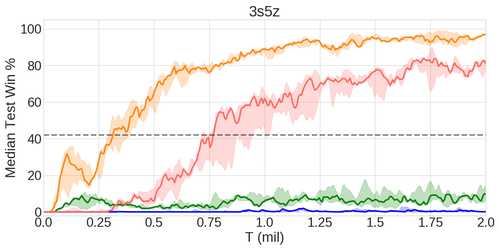}
	\includegraphics[width=0.32\textwidth]{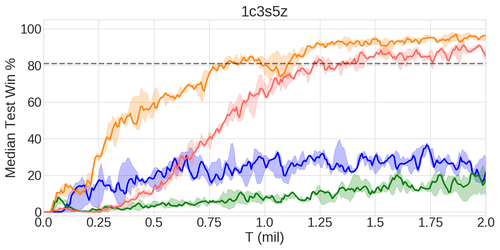}
	\includegraphics[width=0.32\textwidth]{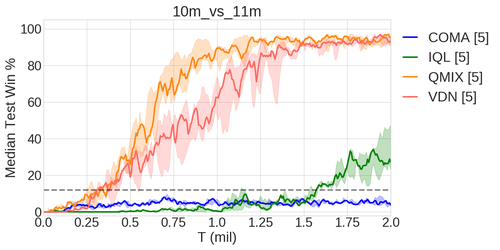}
	\caption{Easy scenarios. The heuristic AI's performance shown as a dotted black line.}
	\label{fig:easy_map_results}
\end{figure*}

Figure \ref{fig:easy_map_results} shows that IQL and COMA struggle even on the \emph{Easy} scenarios, performing poorly on four of the five scenarios in this category.  This shows the advantage of learning a centralised but factored centralised $Q_{tot}$. 
Even though QMIX exceeds $95\%$ test win rate on all of five \emph{Easy} scenarios, they serve an important role in the benchmark as sanity checks when implementing and testing new algorithms. 

\begin{figure*}[h!]
	\centering
	\includegraphics[width=0.32\textwidth]{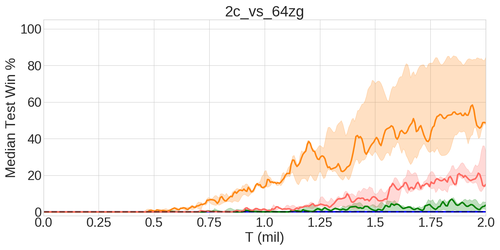}
	\includegraphics[width=0.32\textwidth]{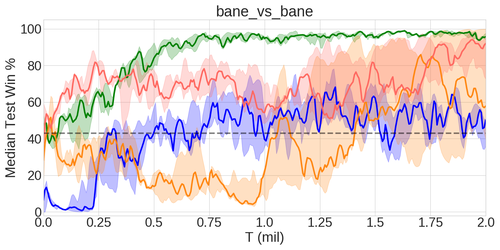}
	\\
	\includegraphics[width=0.32\textwidth]{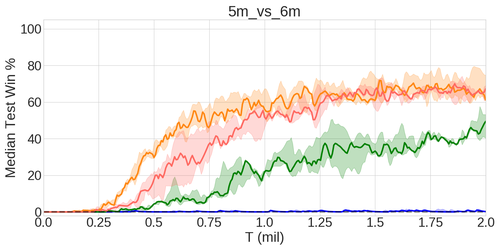}
	\includegraphics[width=0.32\textwidth]{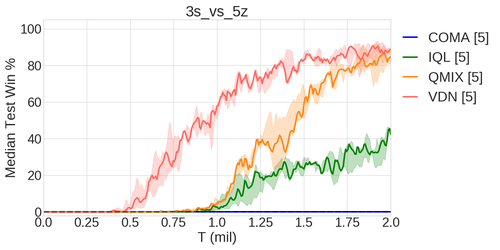}
	\caption{Hard scenarios. The heuristic AI's performance shown as a dotted black line.}
	\label{fig:hard_map_results}
\end{figure*}

The \textit{Hard} scenarios in Figure \ref{fig:hard_map_results} each present their own unique problems.
\textit{2c\_vs\_64zg} only contains 2 allied agents, but 64 enemy units (the largest in the SMAC benchmark) making the action space of the agents much larger than the other scenarios.
\textit{bane\_vs\_bane} contains a large number of allied and enemy units, but the results show that IQL easily finds a winning strategy whereas all other methods struggle and exhibit large variance.
\textit{5m\_vs\_6m} is an asymmetric scenario that requires precise control to win consistently, and in which the best performers (QMIX and VDN) have plateaued in performance.
Finally, \textit{3s\_vs\_5z} requires the three allied stalkers to \textit{kite} the enemy zealots for the majority of the episode (at least 100 timesteps), which leads to a delayed reward problem.

\begin{figure*}[h!]
	\centering
	\includegraphics[width=0.32\textwidth]{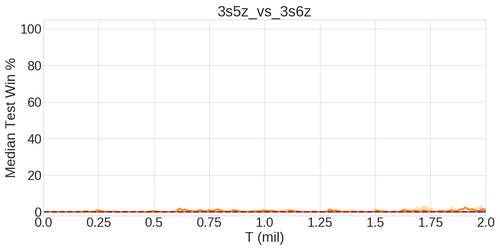}
	\includegraphics[width=0.32\textwidth]{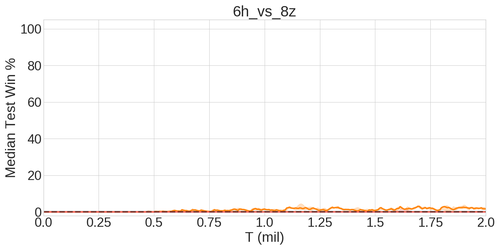}
	\includegraphics[width=0.32\textwidth]{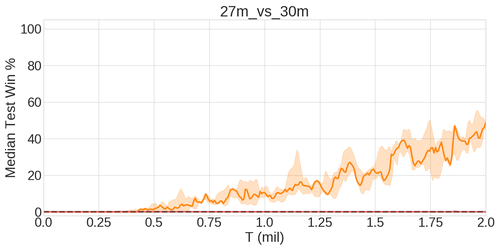}
	\includegraphics[width=0.32\textwidth]{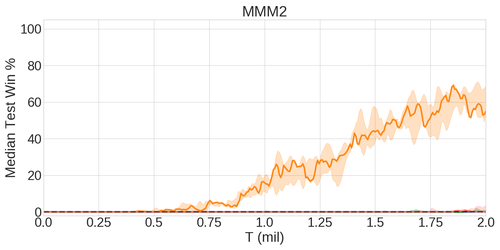}
	\includegraphics[width=0.32\textwidth]{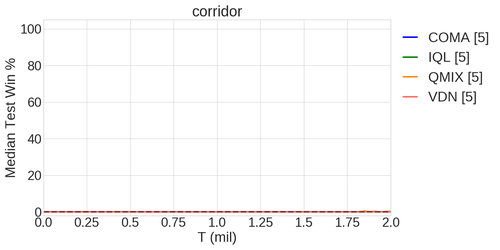}
	\caption{Super Hard scenarios. The heuristic AI's performance shown as a dotted black line.}
	\label{fig:super_hard_map_results}
\end{figure*}

The scenarios shown in Figure \ref{fig:super_hard_map_results} are categorised as \textit{Super Hard} because of the poor performance of all algorithms, with
only QMIX  making meaningful progress on two of the five.
We hypothesise that exploration is a bottleneck in many of these scenarios, providing a nice test-bed for future research in this domain.

\begin{figure*}[h!]
	\centering
	\includegraphics[width=0.32\textwidth]{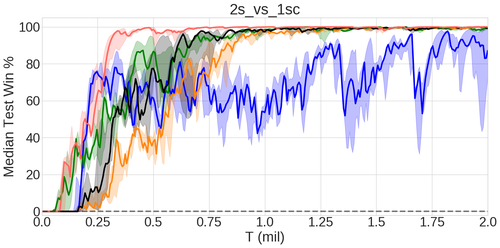}
	\includegraphics[width=0.32\textwidth]{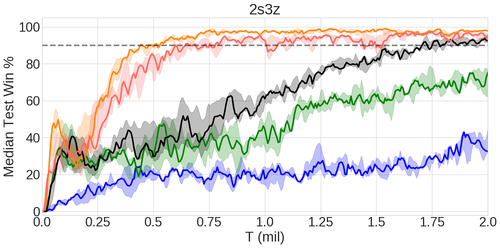}
	\includegraphics[width=0.32\textwidth]{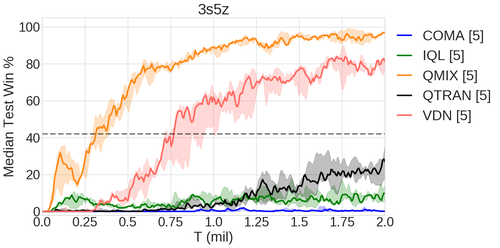}
	\caption{3 scenarios including QTRAN.}
	\label{fig:qtran_results}
		\vspace{-.5cm}
\end{figure*}

We also test QTRAN \cite{son2019qtran} on three of the easiest scenarios, as shown in Figure \ref{fig:qtran_results}. 
QTRAN fails to achieve good performance on \textit{3s5z} and takes far longer to reach the performance of VDN and QMIX on \textit{2s3z}.  
In preliminary experiments, we found the QTRAN-Base algorithm slightly more performant and more stable than QTRAN-Alt. 
For more details on the hyperparameters and architectures considered, please see the Appendix. 

\begin{figure*}[h!]
	\centering
	\includegraphics[width=0.32\textwidth]{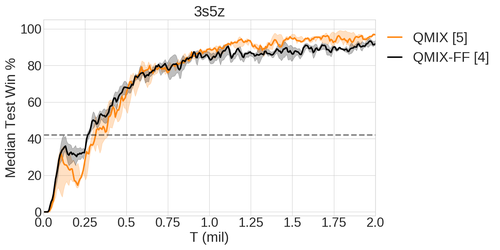}
	\includegraphics[width=0.32\textwidth]{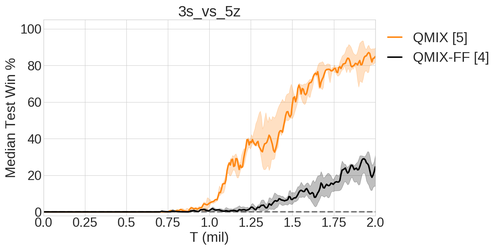}
	\caption{Comparing agent networks with and without RNNs (QMIX-FF) on 2 scenarios.}
	\label{fig:ff_vs_rnn}
	\vspace{-.5cm}
\end{figure*}

Additionally, we compare the necessity of using an agent's action-observation history by comparing the performance of an agent network with and without an RNN in Figure \ref{fig:ff_vs_rnn}.
On the easy scenario of \textit{3s5z} we can see that an RNN is not required to use action-observation information from previous timesteps, but on the harder scenarios of \textit{3s\_vs\_5z} it is crucial to being able to learn how to kite effectively. 

\section{Conclusion and Future Work}
This paper presented SMAC as a set of benchmark problems for cooperative MARL.
Based on the real-time strategy game StarCraft II, SMAC focuses on decentralised
micromanagement tasks and features 14 diverse combat scenarios which challenge
MARL methods to handle partial observability and high-dimensional inputs. We
offer recommendations for reporting evaluations using standardised performance
metrics and provide a thorough report and discussion for several
state-of-the-art MARL algorithms, such as QMIX and COMA. Additionally, we are
open-sourcing PyMARL - our framework for design and analysis of deep MARL
algorithms.

In the near future, we aim to extend SMAC with new challenging scenarios that
feature a more diverse set of units and require a higher level of coordination
amongst agents. Particularly, we plan to make use of the rich skill set of
StarCraft II units, and host scenarios that require the agents to utilise the
features of the terrain. With harder multi-agent coordination problems, we aim to explore the
gaps in existing MARL approaches and motivate further research in this domain,
particularly in areas such as multi-agent exploration and coordination.

We look forward to accepting contributions from the community to SMAC and hope
that it will become a standard benchmark for measuring progress in cooperative
MARL.

\section*{Acknowledgements} 
The authors would like to thank Davit Hayrapetyan for his helpful suggestions about the StarCraft II scenarios. We also thank Phil Bates and Jon Russell (Oracle Corporation) for guidance deploying and executing SMAC on Oracle’s public Cloud Infrastructure and the SC2LE teams at DeepMind and Blizzard for their work on the interface.
	
The work is supported by the European Union's Horizon 2020 research and innovation programme (grant agreement number 637713), the National Institutes of Health (grant agreement number R01GM114311) and EPSRC/MURI grant EP/N019474/1. It is also supported by the an EPSRC grant (EP/M508111/1, EP/N509711/1), the Oxford-Google DeepMind Graduate Scholarship, Microsoft and Free the Drones (FreeD) project under the Innovation Fund Denmark.
The experiments were made possible by generous cloud credit grants from Oracle’s Cloud Innovation Accelerator and NVIDIA.

\bibliography{MultiAgent}
\bibliographystyle{include/icml2019}

\onecolumn
\newpage
\appendix
\section{SMAC}\label{appendix:SMAC}

\subsection{Scenarios}\label{appendix:SMAC_scenarios}

Perhaps the simplest scenarios are \textbf{symmetric} battle scenarios, where the two armies are composed of the same units.
Such challenges are particularly interesting when some of the units are extremely effective against others (this is known as \textit{countering}), for example, by dealing bonus damage to a particular armour type.
In such a setting, allied agents must deduce this property of the game and design an intelligent strategy to protect teammates vulnerable to certain enemy attacks.

SMAC also includes more challenging scenarios, for example, in which the enemy army outnumbers the allied army by one or more units. In such \textbf{asymmetric} scenarios it is essential to consider the health of enemy units in order to effectively target the desired opponent.

Lastly, SMAC offers a set of interesting \textbf{micro-trick} challenges that require a higher-level of cooperation and a specific micro trick to defeat the enemy. 
An example of such scenario is the \texttt{corridor} scenario (Figure~\ref{fig:SC2maps_2}c). Here, six friendly Zealots face 24 enemy Zerglings, which requires agents to make effective use of the terrain features. Specifically, agents should collectively wall off the choke point (the narrow region of the map) to block enemy attacks from different directions. 
The \texttt{3s\_vs\_5z} scenario features three allied Stalkers against five enemy Zealots (Figure~\ref{fig:SC2maps_2}b). Since Zealots counter Stalkers, the only winning strategy for the allied units is to kite the enemy around the map and kill them one after another.
Some of the micro-trick challenges are inspired by \textit{StarCraft Master} challenge missions released by Blizzard \cite{SC2Master}.

\subsection{Environment Setting}\label{appendix:SMAC_evn_setting}

SMAC makes use of the StarCraft II Learning Environment (\textit{SC2LE}) \cite{vinyals_starcraft_2017} to communicate with the StarCraft II engine. SC2LE provides full control of the game by making it possible to send commands and receive observations from the game. However, SMAC is conceptually different from the RL environment of SC2LE. The goal of SC2LE is to learn to play the full game of StarCraft II. This is a competitive task where a centralised RL agent receives RGB pixels as input and performs both macro and micro with the player-level control similar to human players. SMAC, on the other hand, represents a set of cooperative multi-agent micro challenges where each learning agent controls a single military unit.

SMAC uses the \textit{raw API} of SC2LE. Raw API observations do not have any graphical component and include information about the units on the map such as health, location coordinates, etc. The raw API also allows sending action commands to individual units using their unit IDs. This setting differs from how humans play the actual game, but is convenient for designing decentralised multi-agent learning tasks.

Furthermore, to encourage agents to explore interesting micro strategies themselves, we limit the influence of the StarCraft AI on our agents. 
In the game of StarCraft II, whenever an idle unit is under attack, it automatically starts a reply attack towards the attacking enemy units without being explicitly ordered. We disable such automatic replies towards the enemy attacks or enemy units that are located closely by creating new units that are the exact copies of existing ones with two attributes modified: \textit{Combat: Default Acquire Level} is set to \textit{Passive} (default \textit{Offensive}) and \textit{Behaviour: Response} is set to \textit{No Response} (default \textit{Acquire}). These fields are only modified for allied units; enemy units are unchanged.

The sight and shooting range values might differ from the built-in \textit{sight} or \textit{range} attribute of some StarCraft II units. Our goal is not to master the original full StarCraft II game, but rather to benchmark MARL methods for decentralised control.

The game AI is set to level 7, \textit{very difficult}. Our experiments, however, suggest that this setting does significantly impact the unit micromanagement of the built-in heuristics.
\section{Evaluation Methodology}\label{appendix:methodology}

We propose the following methodology for evaluating MARL methods using SMAC. 

To ensure the fairness of the challenge and comparability of results, performances should be evaluated under standardised conditions. 
One should not undertake any changes to the environment used for evaluating the policies. 
This includes the observation and state spaces, action space, the game mechanics, and settings of the environment (e.g., frame-skipping rate).
One should not modify the StarCraft II map files in any way or change the difficulty of the game AI. Episode limits of each scenario should also remain unchanged. %

SMAC restricts the execution of the trained models to be decentralised, i.e., during testing each agent must base its policy solely on its own action-observation history and cannot use the global state or the observations of other agents. %
It is, however, acceptable to train the decentralised policies in centralised fashion. Specifically, agents can exchange individual observations, model parameters and gradients during training as well as make use of the global state. %

\subsection{Evaluation Metrics}

Our main evaluation metric is the mean win percentage of evaluation episodes as a function of environment steps observed, over the course of training. 
Such progress can be estimated by periodically running a fixed number of evaluation episodes (in practice, 32) with any exploratory behaviours disabled. 
Each experiment is repeated using a number of independent training runs and the resulting plots include the median performance as well as the 25-75\% percentiles. 
We use five independent runs for this purpose in order to strike a balance between statistical significance and the computational requirements.
We recommend using the median instead of the mean in order to avoid the effect of any outliers. 
We report the number of independent runs, as well as environment steps used in training. %
Each independent run takes between 8 to 16 hours, depending on the exact scenario, using Nvidia Geforce GTX 1080 Ti graphics cards.

It can prove helpful to other researchers to include the computational resources used, and the wall clock time for running each experiment. SMAC provides functionality for saving StarCraft II replays, which can be viewed using a freely available client. The resulting videos can be used to comment on interesting behaviours observed.

\section{Experimental Setup}\label{appendix:exp_setup}

\subsection{Architecture and Training}

The architecture of all agent networks is a DRQN\cite{hausknecht_deep_2015} with a recurrent layer 
comprised of a GRU with a 64-dimensional hidden state, with a fully-connected 
layer before and after.
Exploration is performed during training using independent $\epsilon$-greedy action selection, where each agent $a$ performs $\epsilon$-greedy action selection over its own $Q_a$. 
Throughout the training, we anneal $\epsilon$ linearly from $1.0$ to $0.05$ over $50k$ time steps and keep it constant for the rest of the learning. 
We set $\gamma = 0.99$ for all experiments.
The replay buffer contains the most recent $5000$ episodes.  
We sample batches of 32 episodes uniformly from the replay buffer, and train on fully unrolled episodes, performing a single gradient descent step after every episode. 
\textbf{Note:} This differs from the earlier an earlier beta release of SMAC, which trained once after every 8 episodes.
The target networks are updated after every $200$ training episodes.

To speed up the learning, we share the parameters of the agent networks across all agents. 
Because of this, a one-hot encoding of the \texttt{agent\_id} is concatenated onto each agent's observations. 
All neural networks are trained using RMSprop\footnote{We set $\alpha = 0.99$ and do not use weight decay or momentum.} with learning rate $5 \times 10^{-4}$. 

The mixing network consists of a single hidden layer of $32$ units, utilising an ELU non-linearity. 
The hypernetworks consist of a feedforward network with a single hidden layer of $64$ units with a ReLU non-linearity\footnote{This differs from the architecture used in \citep{rashid_qmix:_2018} and in an earlier beta release of SMAC, in which a single linear layer was used for the hypernetworks.}. 
The output of the hypernetwork is passed through an absolute function (to acheive non-negativity) and then resized into a matrix of appropriate size.

The architecture of the COMA critic is a feedforward fully-connected neural network with the first 2 layers having 128 units, followed by a final layer of $|U|$ units. We set $\lambda=0.8$. 
We utilise the same $\epsilon$-floor scheme as in \citep{foerster_counterfactual_2017} for the agents’ policies, linearlly annealing $\epsilon$ from 0.5 to 0.01 over 100k timesteps.
For COMA we roll-out 8 episodes and train on those episodes.
The critic is first updated, performing a gradient descent step for each timestep in the episode, starting with the final timestep.
Then the agent policies are updated by a single gradient descent step on the data from all 8 episodes.

The architecture of the centralised $Q$ for QTran is similar to the one used in \citep{son_qtran2019}.
The agent's hidden states ($64$ units) are concatenated with their chosen action ($|U|$ units) and passed through a feedforward network with a single hidden layer and a ReLU non-linearity to produce an agent-action embedding ($64 + |U|$ units). The network is shared across all agents. The embeddings are summed across all agents.
The concatentation of the state and the sum of the embeddings is then passed into the $Q$ network.
The $Q$ network consists of 2 hidden layers with ReLU non-linearities with $64$ units each.
The $V$ network takes the state as input and consists of 2 hidden layers with ReLU non-linearities with $64$ units each.
We set $\lambda_{opt}=1$ and $\lambda_{nopt\_min}=0.1$.

We also compared a COMA style architecture in which the input to $Q$ is the state and the joint-actions encoded as one-hot vectors.
For both architectural variants we also tested having 3 hidden layers.
For both network sizes and architectural variants we performed a hyperparameter search over $\lambda_{opt}=1$ and $\lambda_{nopt\_min} \in \{0.1,1,10\}$ on all 3 of the maps we tested QTran on and picked the best performer out of all configs.

\subsection{Reward and Observation}

All experiments use the default shaped rewards throughout all scenarios. At each timestep, agents receive positive rewards, equal to the hit-point damage dealt, and bonuses of $10$ and $200$ points for killing each enemy unit and winning the scenario, respectively. The rewards are scaled so that the maximum cumulative reward achievable in each scenario is around $20$. 

The agent observations used in the experiments include all features from Section \ref{section:state_and_obs}, except for the he last actions of the allied units (within the sight range), terrain height and walkability.

\section{Table of Results}
\label{sec:table_results}

Table \ref{tab:hai} shows the final median performance (maximum median across the testing intervals within the last $250k$ of training) of the algorithms tested.
The mean test win \%, across 1000 episodes, for the heuristic-based ai is also shown.
\begin{table}[h]
    \setlength{\extrarowheight}{3pt}
        \caption{The Test Win Rate \% of IQL, COMA, VDN, QMIX and the heuristic-based algorithm.}
    \centering
    \begin{center}
        \begin{tabular}{| c | c | c | c | c | c |}
        \hline
        ~&\textbf{IQL}&\textbf{COMA}&\textbf{VDN}&\textbf{QMIX}&\textbf{Heuristic}\\
        \hline \hline
\textbf{2s\_vs\_1sc} &
100&
 98&
100&
100&
0\\
\hline

\textbf{2s3z} &
 75&
 43&
 97&
 99&
90\\
\hline

\textbf{3s5z} &
 10&
  1&
 84&
 97&
42\\
\hline

\textbf{1c3s5z} &
 21&
 31&
 91&
 97&
81\\
\hline

\textbf{10m\_vs\_11m} &
 34&
  7&
 97&
 97&
12\\
\hline
\hline

\textbf{2c\_vs\_64zg} &
  7&
  0&
 21&
 58&
0\\
\hline

\textbf{bane\_vs\_bane} &
 99&
 64&
 94&
 85&
43\\
\hline

\textbf{5m\_vs\_6m} &
 49&
  1&
 70&
 70&
0\\
\hline

\textbf{3s\_vs\_5z} &
 45&
  0&
 91&
 87&
0\\
\hline
\hline

\textbf{3s5z\_vs\_3s6z} &
  0&
  0&
  2&
  2&
0\\
\hline

\textbf{6h\_vs\_8z} &
  0&
  0&
  0&
  3&
0\\
\hline

\textbf{27m\_vs\_30m} &
  0&
  0&
  0&
 49&
0\\
\hline

\textbf{MMM2} &
  0&
  0&
  1&
 69&
0\\
\hline

\textbf{corridor} &
  0&
  0&
  0&
  1&
0\\
\hline
        \end{tabular}
    \end{center}
    \label{tab:hai}
\end{table}

\end{document}